\documentclass[nocompress]{spie}  
\usepackage{pgf}
\usepackage{tikz}
\usepackage[utf8]{inputenc}
\usetikzlibrary{arrows,automata}
\usetikzlibrary{positioning}

\tikzset{
    state/.style={
           rectangle,
           rounded corners,
           draw=black, very thick,
           minimum height=2em,
           inner sep=2pt,
           text centered,
           },
}

\usepackage[]{graphicx}
\usepackage{color}
\usepackage{amssymb}
\usepackage{amsmath}
\usepackage{algorithm,algorithmic}
\usepackage{subcaption}

\newcommand{\blue}[1]{{\color{black} #1}}
\title{\blue{Obstacle Avoidance and Navigation Utilizing Reinforcement Learning with Reward Shaping}} 

\author{Daniel Zhang and Colleen P. Bailey
\skiplinehalf
Department of Electrical Engineering\\ University of North Texas, Denton, TX, USA
}

\authorinfo{Further author information: (Send correspondence to C. P. Bailey.)\\D. Zhang: E-mail: ShengjunZhang@my.unt.edu\\  C. P. Bailey: E-mail: Colleen.Bailey@unt.edu, Telephone: +1 940-891-6874}

\begin{document} 
  \maketitle 

\begin{abstract}

\blue{
In this paper, we investigate the obstacle avoidance and navigation problem in the robotic control area. For solving such a problem, we propose revised Deep Deterministic Policy Gradient (DDPG) and Proximal Policy Optimization algorithms with an improved reward shaping technique. We compare the performance between the original DDPG and PPO with the revised version of both on simulations with a real mobile robot and demonstrate that the proposed algorithms achieve better results.
}

\end{abstract}



\section{INTRODUCTION}
\label{sec:intro}  

Unmanned Ground Vehicles (UGVs) and Unmanned Aerial Vehicles (UAVs) are widely used in both civil and military applications. In the civil domain, unmanned vehicles are used for aerial crop surveys, search and rescue, inspection of power lines and pipelines and more. 
Using UGVs and UAVs for military-based scenarios has multiple benefits including reducing the risk of death by replacing human operators. In this paper, we mainly focus on UGVs.

\blue{
Navigating UGVs without being trapped by obstacles is an essential problem in both academia and industry.
Traditionally, for such a problem, only simultaneous localization and mapping (SLAM) techniques are adopted. It is difficult to apply a single SLAM algorithm or scheme for all different types of environments.
Recently, with the success of reinforcement learning (RL) in many applications, it has gained more and more attraction in research.
It is natural to use RL to help the autonomous agents, in this case, UGVs, make decisions in complex environments instead of SLAM.
}

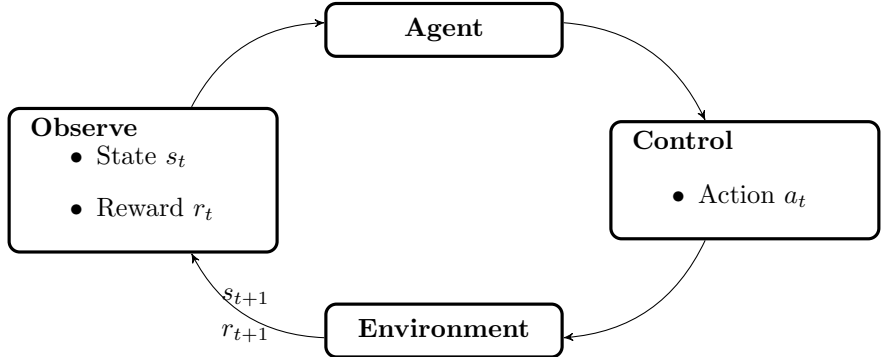
\begin{figure}[ht]

\centering
\begin{tikzpicture}[->,>=stealth']
 \node[state] (Observe) 
 {\begin{tabular}{l}
  \textbf{Observe}\\
  \parbox{3cm}{\begin{itemize}
   \item State $s_t$
   \item Reward $r_t$
  \end{itemize}
  }
 \end{tabular}};
  
 \node[state,    	
  text width=3cm, 	
  yshift=2cm, 		
  right of=Observe, 	
  node distance=4cm, 	
  anchor=center] (Agent) 	
 {%
 \begin{tabular}{l} 	
  \textbf{Agent}
 \end{tabular}
 };
 
 \node[state,
  below of=Agent,
  yshift=-3cm,
  anchor=center,
  text width=3cm] (Environment) 
 {%
 \begin{tabular}{l}
  \textbf{Environment}
 \end{tabular}
 };

 \node[state,
  right of=Agent,
  yshift=-2cm,
  node distance=4cm,
  anchor=center] (Control) 
 {%
 \begin{tabular}{l}
 \textbf{Control}\\
  \parbox{3cm}{\begin{itemize}
   \item Action $a_t$
  \end{itemize}}
 \end{tabular}
 };

 \path (Observe) 	edge[bend left=30]  node[anchor=south,above]{}
                                    node[anchor=north,below]{} (Agent)
 (Environment)     	edge[bend left=30] node[anchor=south,above]{$s_{t+1}$}node[anchor=north,below]{$r_{t+1}$}(Observe)
 (Agent)       	edge                                        [bend left=30](Control)
 (Control)       	edge[bend left]                                          (Environment)
%
;

\end{tikzpicture}\caption[rl] 
{ \label{fig:rl} 
A typical reinforcement learning flow       \cite{sutton2018reinforcement} .}
\end{figure}

Reinforcement learning is an approach that helps an agent learn how to \blue{make optimal decisions} from the environment. 
\blue{In general}, reinforcement learning is modeled as a Markov Decision Process (MDP)\cite{zhang2018fully} \blue{as shown in} Fig.~\ref{fig:rl}. Typically, the RL agent \blue{takes} an action $a_t$ following a policy $\pi$ based on the observation of the state $s_t$ and reward $r_t$ at time $t$.
Since the action $a_t$ is applied in the environment by the agent, the new state changes to $s_{t+1}$ and a reward $r_{t+1}$ is assigned to the agent. 

\blue{With} the growth of \blue{complex environments,} such as \blue{UGV applications}, combining deep learning and reinforcement learning in continuous action space control has attracted \blue{increasing} attention, specifically \blue{in} the robotic control area.
Deep Q-Learning (DQN) has been applied in robotic control successfully, however, Q-Learning is designed to  deal  with  the  discrete-time  decision-making  problem. 
Deep Deterministic Policy Gradient (DDPG)\cite{lillicrap2015continuous} methods have obtained popularity due to successful simulated applications in various robotic control tasks. However, DDPG has been benchmarked on real-world robots and performs more poorly than Proximal Policy Optimization (PPO) \cite{schulman2017proximal, mahmood2018benchmarking}.
Additionally, a new reward shaping technique\cite{dong2020principled} has been applied in several simulations in OpenAI Gym\cite{openaigym} and was shown to improve the performance significantly in terms of accumulated reward. However, these simulations are not complex enough in reality.

In this paper, we apply \blue{both DDPG and PPO} algorithms equipped with the new reward shaping technique\cite{dong2020principled} \blue{in an obstacle avoidance robotic control problem.} 
Gazebo, ROS, and Turtlebot 3 Burger\textsuperscript{\textregistered} are used as a platform to demonstrate \blue{the proposed algorithms} and compare the performances \blue{with/without the improved reward shaping technique} when applied to the same real mobile robotic control problem.


\section{Algorithms}

\subsection{State-of-the-art of Reinforcement Learning}

There are two main approaches to solving RL problems: methods based on \textit{value functions}, for instance DQN, and methods based on \textit{policy search}, such as REINFORCE\cite{williams1992simple}, PPO and DDPG. 

Value function based methods learn the optimal policy indirectly from value functions. 
For example, in $Q$-learning, the agent learns from the state-action value function, known as the $Q$-value, and updates the optimal $Q$-value in a $Q$-table as the optimal policy. When the agent completes training, the actions are chosen from the $Q$-table. 
The drawbacks of $Q$-learning include a $Q$-table exponentially exploding when handling complex tasks. Additionally, $Q$-learning is not suitable for continuous action tasks. 
Deep $Q$-learning solves the huge $Q$-table issue by embedding a neural network, however, it still suffers in continuous action tasks.

Policy search methods directly search for an optimal policy $\pi^{\ast}$.
Typically, a parameterized policy $\pi_{\theta}$ is chosen, whose parameters are updated to maximize the expected reward $\mathbb{E}[R|\theta]$ using either gradient-based or gradient-free optimization methods.

\subsection{Deep Deterministic Policy Gradient}
\label{sec:ddpg}
Deep Deterministic Policy Gradient\cite{lillicrap2015continuous} is a policy-gradient actor-critic algorithm, which combines Deterministic Policy Gradient(DPG)\cite{silver2014deterministic} and DQN \cite{mnih2013playing}, as summarized in Algorithm~\ref{alg:ddpg}. 
By applying the actor-critic framework while learning a deterministic policy, DDPG is able to solve continuous space learning tasks. 
An actor network is used to optimize the parameter $\theta$ for the policy and a critic network evaluates the policy generated or optimized in the actor network based on temporal difference (TD).
In DDPG, the two target networks are initialized at the start of training, which save the copies of the state-action value function $Q(s, a)$.

DDPG is a breakthrough that enables agents to choose actions in a continuous space and perform well.
The main drawback of DDPG is the difficulty of choosing the appropriate step size.
A small step size leads to a slow convergence rate, while a large one tends to affect the sampling from the replay buffer and the estimators of the value function, so the policy improvement is not guaranteed and gives a really poor performance.

\begin{algorithm}[ht]
\caption{DDPG}
\label{alg:ddpg}
\begin{algorithmic}[1]
\STATE \textbf{Initialization}: Randomly initialize critic network $Q$ and actor network $\mu$.
\STATE \textbf{Initialization}: Initialize target network $Q\prime$ and $\mu\prime$.
\STATE \textbf{Initialization}: Initialize replay buffer $B$.
\FOR{episode$=0,1,\dots$}
\STATE \textbf{Initialization}: Initialize a random process for action exploration.
\STATE Observe state $s_1$.
\FOR{iteration$=1,\dots,T$}
\STATE  Select action $a_t$ based on current policy, execute $a_t$ and observe both reward $r_t$ and state $s_{t+1}$;
\STATE  Store $(s_t, a_t, r_t, s_{t+1})$ in buffer $B$;
\STATE Sample a minibatch of N samples from $B$;
\STATE Update critic network $Q$;
\STATE Update actor network $\mu$;
\STATE Update target networks $Q\prime$ and $\mu\prime$ simultaneously;
\ENDFOR
\ENDFOR
\end{algorithmic}
\end{algorithm}

\subsection{Proximal Policy Optimization} 
\label{sec:ppo}
Before PPO, in order to guarantee policy improvement, Trust Region Policy Optimization (TRPO) \cite{schulman2015trust} introduced $\mathcal{KL}$ divergence to measure whether the new policy is better than the average performance of the old policy as an optimization constraint.
With the $\mathcal{KL}$ divergence constraint, the policy is guaranteed to improve monotonically.
However, TRPO is difficult to implement and requires more computation to execute.

Proximal Policy Optimization (PPO)\cite{schulman2017proximal} proposed a clipped surrogate objective function that reduces the computation from the constrained optimization.
The loss function in TRPO is given by:
\begin{equation}\label{eq:trpoloss}
L(\theta) = \hat{\mathbb{E}}_{t}[r_t(\theta)\hat{A_{t}}]
\end{equation}
where $\hat{\mathbb{E}}_{t}[\dots]$  denotes the empirical average over a finite batch of samples, $\hat{A_{t}}$ is the estimator of advantage function $\hat{A_{t}}:= -V(s_{t})+r_{t}+\gamma r_{t+1} +\dots+\gamma^{T-r}V(s_{T})$, and $r_t(\theta)$ denotes the probability ratio between current policy and old policy $r_t(\theta) := \frac{\pi_{\theta}(a_{t}|s_{t})}{\pi_{\theta_{old}}(a_{t}|s_{t})}$.

The $\mathcal{KL}$ divergence constraint forbids a drastic update from the old policy to the new policy, PPO applies a penalty to avoid such a huge change. The clipped surrogate objective function is given by:
\begin{equation}\label{eq:ppoloss}
L^{CLIP}(\theta) = \hat{\mathbb{E}}_{t}[\min(r_t(\theta)\hat{A}_{t}, \text{clip}(r_t(\theta), 1-\epsilon, 1+\epsilon)\hat{A}_{t} )]
\end{equation}
Compared to TRPO, the probability ratio $r_t(\theta)$ is clipped between $[1-\epsilon, 1+\epsilon]$, in practice, we choose $\epsilon = 0.2$, which means no matter how good the new policy, the $r_t(\theta)$ only increases 20$\%$ at most.

$\hat{A}_{t}  \geq 0$ means the current action performs better than others under a specific state. If the new policy is better than the old one, $r_t(\theta)$ should be increased so that the better action has a higher probability to be chosen. In contrast, for $\hat{A}_{t}  \leq 0$, the action should be discouraged and $r_t(\theta)$ should be decreased. 

Note that the loss function of PPO in Eq.~\ref{eq:ppoloss} is the lower bound of Eq.~\ref{eq:trpoloss}. Also, the computation is reduced due to the KL divergence constraint. 
The PPO algorithm is summarized	 as Algorithm~\ref{alg:ppo}.

\begin{algorithm}[ht]
\caption{PPO}
\label{alg:ppo}
\begin{algorithmic}[1]
\FOR{episode$=0,1,\dots$}
\FOR{iteration$=1,\dots,N$}
\STATE  Run policy $\pi_{old}$ in environment for $T$ timesteps;
\STATE  Compute advantage estimates $\hat{A}_{1}, \dots, \hat{A}_{T}$;
\ENDFOR
\STATE Optimize $L^{CLIP}(\theta)$ with respect to $\theta$, with $K$ epochs and minibatch size $M \leq NT$.
\STATE Update $\theta_{old} \leftarrow \theta$
\ENDFOR
\end{algorithmic}
\end{algorithm}

\subsection{\blue{Proposed Algorithms with Reward Shaping}}

The drawbacks of reinforcement learning include long convergence time, enormous training data size, and difficult reproduction.
A reward shaping technique based on the Lyapunov stability theory~\cite{dong2020principled} accelerates the convergence of RL algorithms.
Inspired by such a technique, we implement the reward shaping method in Eq.~\ref{eq:reward}
\begin{equation}\label{eq:reward}
R^{lyap}(s_{t+1}, a_{t+1}) = R(s_{t},a_{t}) + \eta( \gamma R(s_{t+1}, a_{t+1}) - R(s_{t},a_{t}) )
\end{equation}
where $\eta$ is a tuning parameter that weights the shaped term $ \gamma R(s_{t+1}, a_{t+1}) - R(s,a)$. 
The reward shaping in Eq.~\ref{eq:reward} has shown to guarantee convergence, preserve optimality and lead to an unbiased optimal policy.

\blue{
In DDPG, we adopt the reward shaping technique in the actor network based on the TD error, recall TD error $:= r_{t+1} +\gamma V(s_{t+1}) -V(s_t)$. After equipping with reward shaping, the new TD error is given in Eq.~\ref{eq:TD}
\begin{align}\label{eq:TD}
\text{TD error:} = r_{t+1}+\gamma \mathbb{E}[R^{lyap}(s_{t+1}, a_{t+1})] - \mathbb{E}[R^{lyap}(s_{t}, a_{t})]
\end{align}
In PPO, the reward shaping is applied to the estimator of advantage function $\hat{A}_{t}$, which is given in Eq.~\ref{eq:ad}
\begin{equation}\label{eq:ad}
\hat{A}_{t} = -\mathbb{E}[R^{lyap}(s_t,a_t)]+r_{t}+\gamma r_{t+1}+\dots+\gamma^{T-r}\mathbb{E}[R^{lyap}(s_T, a_T)]
\end{equation}
}
\section{Simulation}
\label{sec:sim}
\begin{figure}[ht!]
\begin{center}
\includegraphics[height=5cm]{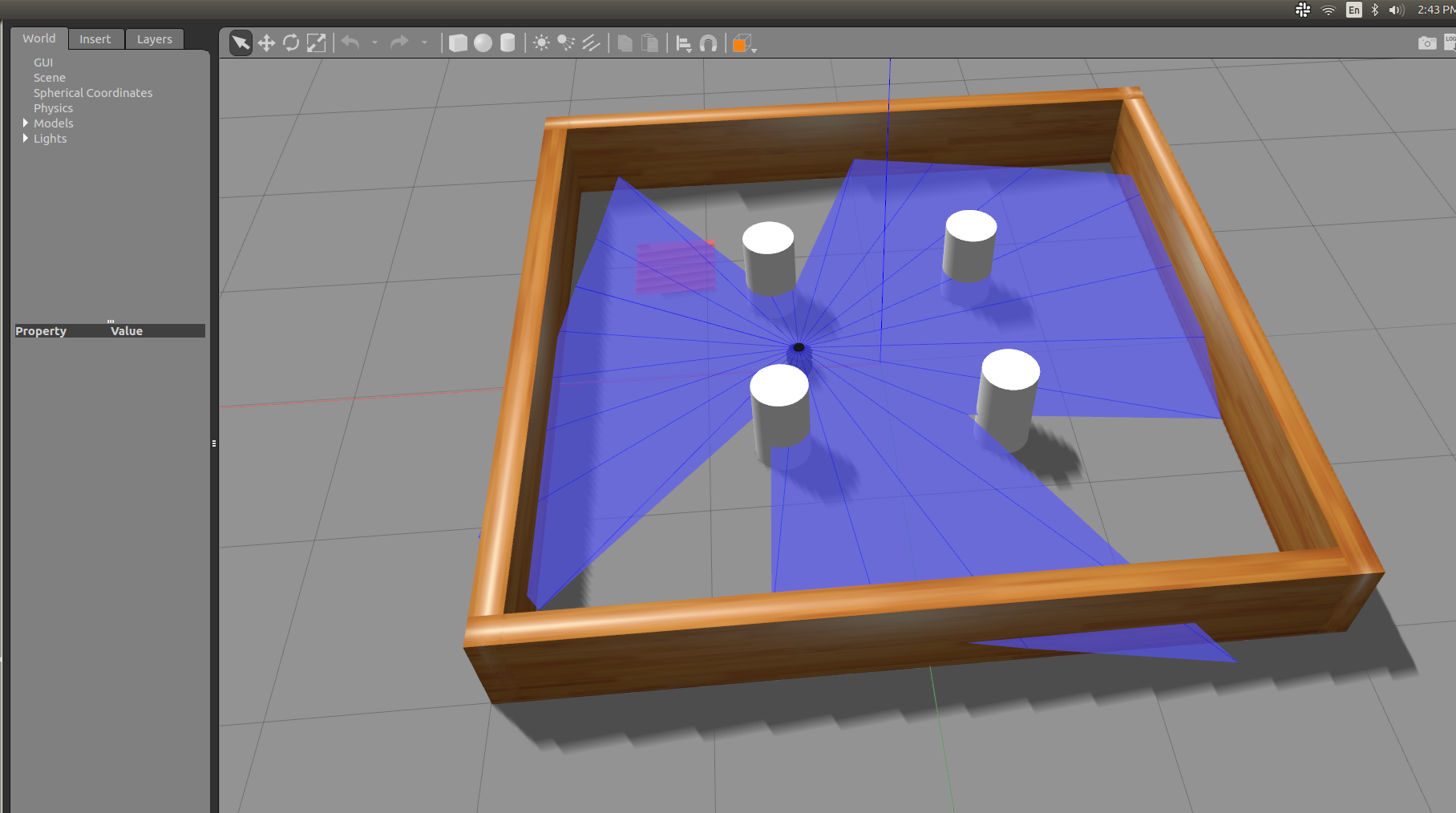}
\end{center}
\caption[Setup exp] 
{ \label{fig:exp} 
Gazebo environment for training.}
\end{figure}

All the programs are conducted in \texttt{Python}, running on a computer node with \texttt{Intel Core i5-9600K processor, Nvidia RTX 2070 super, 32~GB RAM, Ubuntu 16.04}. We use Gazebo, ROS, and Turtlebot 3 Burger\textsuperscript{\textregistered} to demonstrate both DDPG and PPO separately. 
The training environment \blue{set up for demonstrating the obstacle avoidance and navigation task in Gazebo is shown in Fig~\ref{fig:exp}}.
The 4 white cylinders are the obstacles, the red square is the target for the robot, and the blue lines demonstrate the LiDAR scanning from the robot. \blue{Training details are given in Table~\ref{tab:details}.} The state dimension is 26, which contains 24 LiDAR values. The action dimension is 2; one is changing the linear velocity, while the other is changing the angular velocity. The minibatch size we choose here is 32 and the Optimizer is Adam \cite{kingma2014adam}, with learning rate 0.0003, $\beta_1 = 0.9$, and $\beta_2 = 0.999$. 

\begin{table}[ht!]
\caption{Training details.} 
\label{tab:details}
\begin{center}
\scalebox{1.0}{      
\begin{tabular}{|l|l|l|} 
\hline
\rule[-1ex]{0pt}{3.5ex}   & DDPG &  PPO \\
\hline
\rule[-1ex]{0pt}{3.5ex}  Episode & 1000 &  1000 \\
\hline
\rule[-1ex]{0pt}{3.5ex}  State dimension & 26 & 26 \\
\hline
\rule[-1ex]{0pt}{3.5ex}  Action dimension & 2 &  2  \\
\hline
\rule[-1ex]{0pt}{3.5ex}  Batch size & 32 & 32  \\
\hline
\rule[-1ex]{0pt}{3.5ex}  Optimizer & Adam & Adam  \\
\hline
\rule[-1ex]{0pt}{3.5ex}  Reward shaping parameter $\eta$ & 0.4 & 0.4  \\
\hline
\end{tabular}}
\end{center}
\end{table} 

\begin{figure*}[ht]
    \centering
    \begin{subfigure}[t]{0.5\textwidth}
        \centering
        \includegraphics[height=5cm]{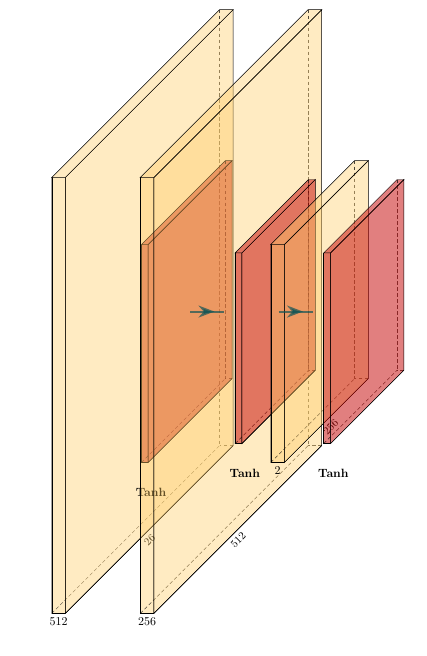}
        \caption[actor] 
{ \label{fig:actor} 
PPO actor network.}
    \end{subfigure}%
    ~ 
    \begin{subfigure}[t]{0.5\textwidth}
        \centering
        \includegraphics[height=5cm]{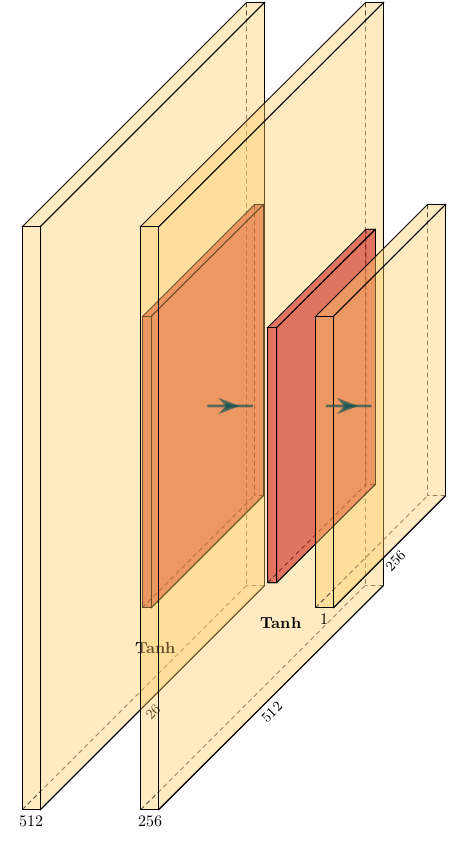}
        \caption[critic] 
{ \label{fig:critic} 
PPO critic network.}
    \end{subfigure}
    \caption{\label{fig:nnppo} Neural Network model.}
\end{figure*}

In Fig.~\ref{fig:nnppo}, we demonstrate the \blue{neural network chosen} for the PPO algorithm in actor-critic style.

\section{results}
\label{sec:results}
\begin{figure*}[ht!]
    \centering
    \begin{subfigure}[t]{0.5\textwidth}
        \centering
        \includegraphics[height=6cm]{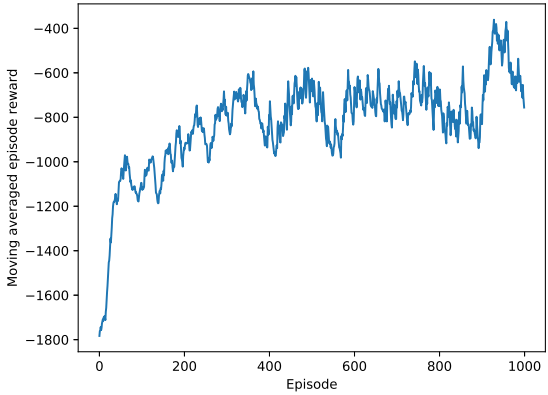}
        \caption[ddpgog] 
{ \label{fig:ddpg_og} 
DDPG reward w/o RS.}
    \end{subfigure}%
    ~ 
    \begin{subfigure}[t]{0.5\textwidth}
        \centering
        \includegraphics[height=6cm]{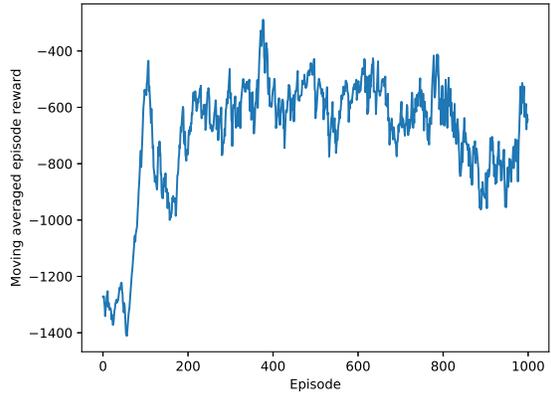}
        \caption[ppoog] 
{ \label{fig:ppo_og} 
PPO reward w/o RS.}
    \end{subfigure}
    \caption{\label{fig:performance_og} Comparison of performance without reward shaping.}
\end{figure*}

\begin{figure*}[ht!]
    \centering
    \begin{subfigure}[t]{0.5\textwidth}
        \centering
        \includegraphics[height=6cm]{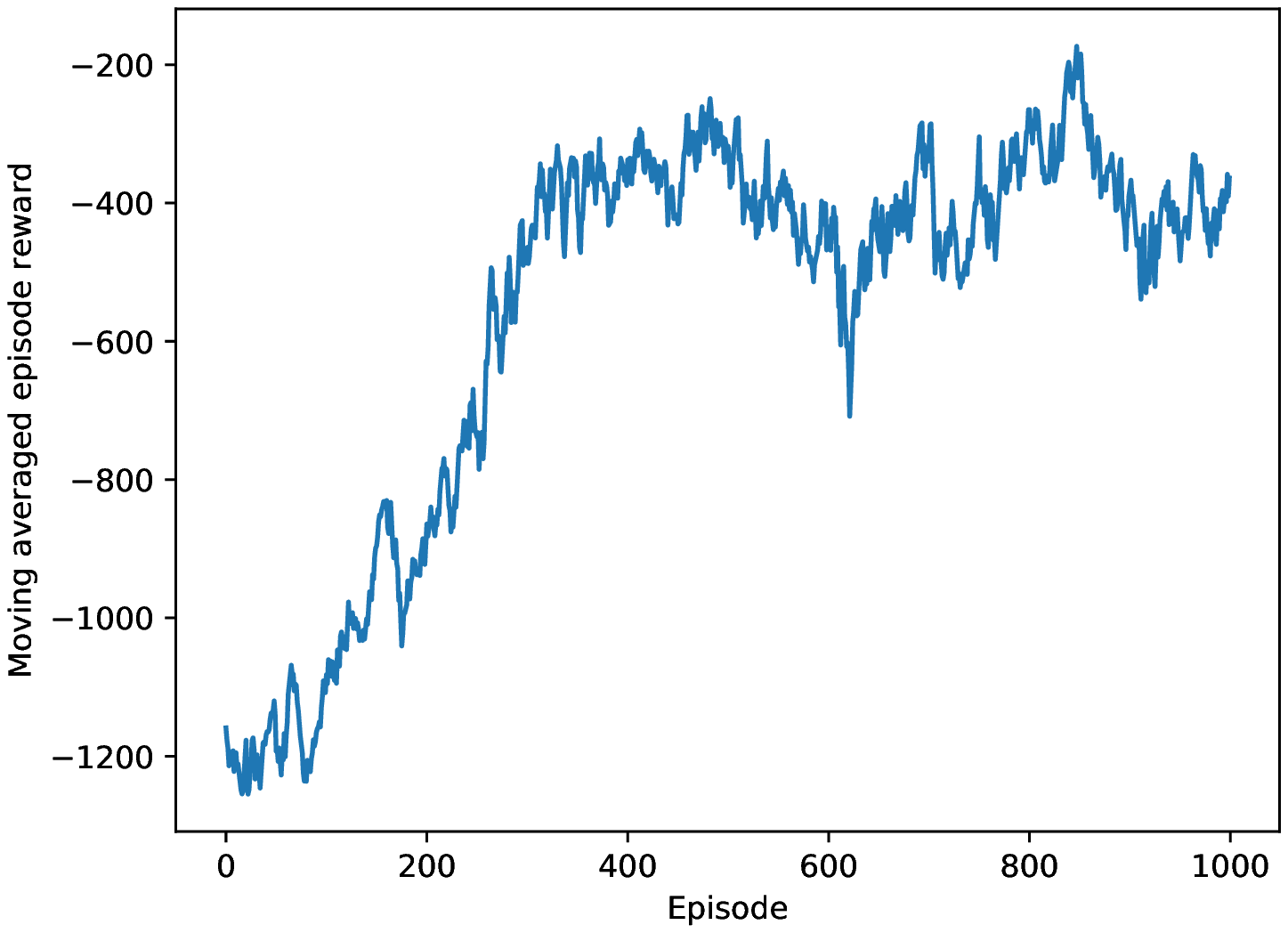}
        \caption[ddpg] 
{ \label{fig:ddpg} 
DDPG reward with RS.}
    \end{subfigure}%
    ~ 
    \begin{subfigure}[t]{0.5\textwidth}
        \centering
        \includegraphics[height=6cm]{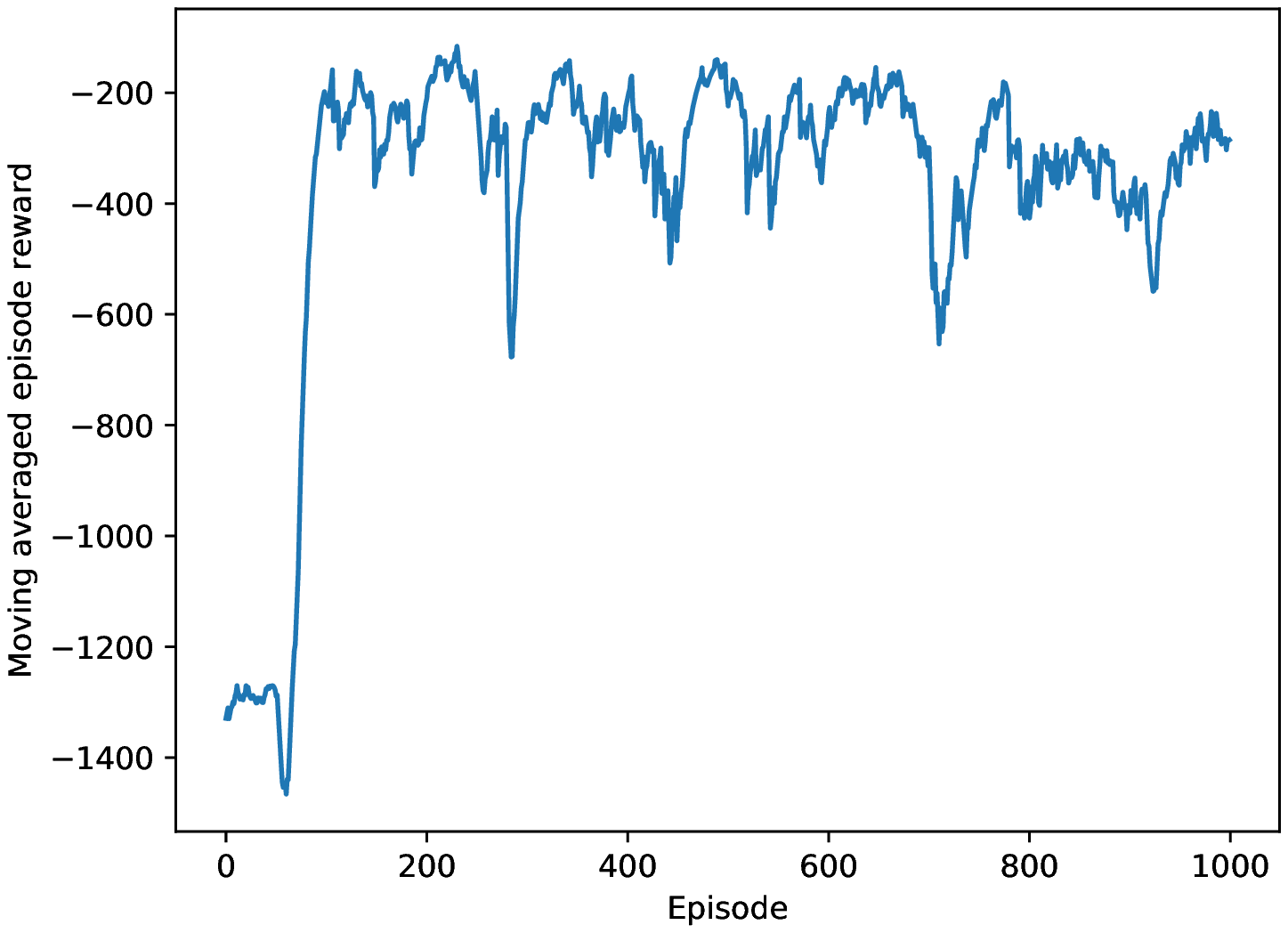}
        \caption[ppo] 
{ \label{fig:ppo} 
PPO reward with RS.}
    \end{subfigure}
    \caption{\label{fig:performance} Comparison of performance with reward shaping.}
\end{figure*}

Compareing the two algorithms, PPO is more implementation friendly than DDPG.	  
Fig.~\ref{fig:ddpg_og} and Fig.~\ref{fig:ppo_og} \blue{show the averaged rewards values in DDPG and PPO without any reward shaping technique.}
\blue{In Fig.~\ref{fig:ddpg} and Fig.~\ref{fig:ppo}, the averaged reward values are shown for both DDPG and PPO algorithms with the improved reward shaping technique. }

We divide the comparison of performance into two parts.
First we compare DDPG and PPO, then we compare the effectiveness of reward shaping technique.

Comparing~\ref{fig:ddpg_og} with~\ref{fig:ppo_og} and~\ref{fig:ddpg} with~\ref{fig:ppo} respectively, we see that PPO converges faster than DDPG and has better averaged reward values.
Comparing~\ref{fig:ddpg_og} with~\ref{fig:ddpg} and~\ref{fig:ppo_og} with~\ref{fig:ppo} separately, both DDPG and PPO with the reward shaping technique achieve a better performance than the original version.
The experimental results are listed statistically in Table~\ref{tab:result}, where the minimum, maximum and average rewards are provided corresponding to each learning algorithm.

\begin{table}[ht!]
\caption{Result details.} 
\label{tab:result}
\begin{center}
\scalebox{1.0}{      
\begin{tabular}{|c|c|c|c|} 
\hline
\rule[-1ex]{0pt}{3.5ex}   & min &  max & avg\\
\hline
\rule[-1ex]{0pt}{3.5ex}  DDPG w/o reward shaping & -1790.45 &  -360.87& -710.56\\
\hline
\rule[-1ex]{0pt}{3.5ex}  PPO w/o reward shaping & -1407.45 & -298.78& -679.43\\
\hline
\rule[-1ex]{0pt}{3.5ex}  DDPG with reward shaping & \textbf{-1290.22} &  -177.65 & -477.06\\
\hline
\rule[-1ex]{0pt}{3.5ex} PPO with reward shaping & -1478.23 & \textbf{-155.06}  & \textbf{-323.59}\\
\hline
\end{tabular}}
\end{center}
\end{table}

\section{Conclusion}

In this paper, we consider solving the obstacle avoidance and navigation problem for unmanned ground vehicles by applying DDPG and PPO equipped with a reward shaping technique. 
\blue{We compare DDPG and PPO in the same learning settings.
The simulations show that PPO has a better performance than DDPG and the proposed algorithms help RL achieve better results.
For future directions, we will investigate PPO applied to multi-agent robots systems and combine SLAM techniques and reinforcement learning to improve the performance.}

\bibliography{report}   

\begin{thebibliography}{10}

\bibitem{zhang2018fully}
Zhang, K., Yang, Z., Liu, H., Zhang, T., and Ba{\c{s}}ar, T., ``Fully
  decentralized multi-agent reinforcement learning with networked agents,''
  {\em arXiv preprint arXiv:1802.08757}  (2018).

\bibitem{sutton2018reinforcement}
Sutton, R.~S. and Barto, A.~G.,  [{\em Reinforcement learning: An
  introduction}{\nolinebreak\hspace{0.1em}]} (2018).

\bibitem{lillicrap2015continuous}
Lillicrap, T.~P., Hunt, J.~J., Pritzel, A., Heess, N., Erez, T., Tassa, Y.,
  Silver, D., and Wierstra, D., ``Continuous control with deep reinforcement
  learning,'' {\em arXiv preprint arXiv:1509.02971}  (2015).

\bibitem{schulman2017proximal}
Schulman, J., Wolski, F., Dhariwal, P., Radford, A., and Klimov, O., ``Proximal
  policy optimization algorithms,'' {\em arXiv preprint arXiv:1707.06347}
  (2017).

\bibitem{mahmood2018benchmarking}
Mahmood, A.~R., Korenkevych, D., Vasan, G., Ma, W., and Bergstra, J.,
  ``Benchmarking reinforcement learning algorithms on real-world robots,'' in
  [{\em Conference on Robot Learning}{\nolinebreak\hspace{0.1em}]},   561--591
  (2018).

\bibitem{dong2020principled}
Dong, Y., Tang, X., and Yuan, Y., ``Principled reward shaping for reinforcement
  learning via lyapunov stability theory,'' {\em Neurocomputing}  (2020).

\bibitem{openaigym}
Brockman, G., Cheung, V., Pettersson, L., Schneider, J., Schulman, J., Tang,
  J., and Zaremba, W., ``Openai gym,'' (2016).

\bibitem{williams1992simple}
Williams, R.~J., ``Simple statistical gradient-following algorithms for
  connectionist reinforcement learning,'' {\em Machine learning}~{\bf 8}(3-4),
  229--256 (1992).

\bibitem{silver2014deterministic}
Silver, D., Lever, G., Heess, N., Degris, T., Wierstra, D., and Riedmiller, M.,
  ``Deterministic policy gradient algorithms,'' (2014).

\bibitem{mnih2013playing}
Mnih, V., Kavukcuoglu, K., Silver, D., Graves, A., Antonoglou, I., Wierstra,
  D., and Riedmiller, M., ``Playing atari with deep reinforcement learning,''
  {\em arXiv preprint arXiv:1312.5602}  (2013).

\bibitem{schulman2015trust}
Schulman, J., Levine, S., Abbeel, P., Jordan, M., and Moritz, P., ``Trust
  region policy optimization,'' in [{\em International conference on machine
  learning}{\nolinebreak\hspace{0.1em}]},   1889--1897 (2015).

\bibitem{kingma2014adam}
Kingma, D.~P. and Ba, J., ``Adam: A method for stochastic optimization,'' {\em
  arXiv preprint arXiv:1412.6980}  (2014).

\end{thebibliography}
\bibliographystyle{spiebib}   
\end{document}